%% file: main.tex
\documentclass[letterpaper, 10 pt, conference]{ieeeconf}  % Comment this line out if you need a4paper

\IEEEoverridecommandlockouts                              % This command is only needed if 
\usepackage{graphics} % for pdf, bitmapped graphics files
\usepackage{epsfig} % for postscript graphics files
\usepackage{times} % assumes new font selection scheme installed
\usepackage{amsmath} % assumes amsmath package installed
\usepackage{amssymb}  % assumes amsmath package installed
\usepackage[bookmarks=true]{hyperref}
\usepackage{xcolor}
\usepackage{color}
\usepackage[algo2e]{algorithm2e} 
\usepackage{multirow}
\usepackage{bm}

\usepackage{algorithm}
\usepackage{algorithmic}

\definecolor{Gray}{gray}{0.85}
\definecolor{LightCyan}{rgb}{0.88,1,1}
\definecolor{somegray}{rgb}{0.5, 0.5, 0.5}
\newcommand{\darkgrayed}[1]{\textcolor{somegray}{#1}}
\makeatletter

\newcommand*\titleheader[1]{\gdef\@titleheader{#1}}
\AtBeginDocument{%
  \let\st@red@title\@title
  \def\@title{%
    \vskip-2.8em
    \bgroup\normalfont\large\centering\@titleheader\par\egroup
    \vskip0.9em\st@red@title}
}
\makeatother

\titleheader{\darkgrayed{This paper has been accepted for publication at the
\\ IEEE International Conference on Robotics and Automation (ICRA), Xi’an, 2021. 
\copyright IEEE} }

\title{\LARGE \bf
No Need for Interactions: Robust Model-Based \\
Imitation Learning using Neural ODE 
}

\author{
    HaoChih Lin$^{1, 2}$,
    Baopu Li$^{1*}$,
    Xin Zhou$^{1*}$, Jiankun Wang$^{3}$, and Max Q.-H. Meng $^{3,4}$ \textit{Fellow IEEE}
    
    \thanks{* means corresponding authors. $^{1}$The authors are with the Baidu Research(USA). 
    $^{2}$The author is a master student at ETH Zurich, Switzerland.
    $^{3}$The authors are with the Department of Electronic and Electrical Engineering of the Southern University of Science and Technology, Shenzhen, China.
    $^4$The author is on temporary leave from the Department of Electronic Engineering, The Chinese University of Hong Kong, Hong Kong.}
    
}

\begin{document}

\maketitle

\input{sections/abstract}
\input{sections/introduction}

\input{sections/preliminary}
\input{sections/methodology}
\input{sections/experiments}
\input{sections/conclusion}

%%%%%%%%%%%%%%%%%%%%%%%%%%%%%%%%%%%%%%%%%%%%%%%%%%%%%%%%%%%%%%%%%%%%%%%%%%%%%%%%
\bibliographystyle{IEEEtran}
\bibliography{ref}

\end{document}

%% file: sections/abstract.tex
\begin{abstract}
Interactions with either environments or expert policies during training are needed for most of the current imitation learning (IL) algorithms. For IL problems with no interactions, a typical approach is Behavior Cloning (BC). However, BC-like methods tend to be affected by distribution shift. To mitigate this problem, we come up with a Robust Model-Based Imitation Learning (RMBIL) framework that casts imitation learning as an end-to-end differentiable nonlinear closed-loop tracking problem. RMBIL applies Neural ODE to learn a precise multi-step dynamics and a robust tracking controller via Nonlinear Dynamics Inversion (NDI) algorithm. Then, the learned NDI controller will be combined with a trajectory generator, a conditional VAE, to imitate an expert's behavior. Theoretical derivation shows that the controller network can approximate an NDI when minimizing the training loss of Neural ODE. Experiments on Mujoco tasks also demonstrate that RMBIL is competitive to the state-of-the-art generative adversarial method (GAIL) and achieves at least 30\% performance gain over BC in uneven surfaces. % environments.
\end{abstract}

%% file: sections/introduction.tex
\section{INTRODUCTION}
\vspace{-6pt}
For the majority of the recent Imitation Learning (IL) works, interactions with environments are considered, 
%In such a scenario, 
and several algorithms have been proposed to solve the IL problem in this context, such as Inverse Reinforcement Learning (IRL)  \cite{ng2000algorithms, abbeel2004apprenticeship}. % which attempts to recover the cost function under which the expert is optimal. 
The state-of-the-art Generative Adversarial Imitation Learning (GAIL) \cite{ho2016generative} %, which learns an optimal policy by performing occupancy measure matching, 
is also based on prior IRL works. %Although 
The success of GAIL gains popularity of the adversarial IL (AIL) framework %to become the major trend 
in IL research field \cite{ merel2017learning, ding2019goal, wu2019imitation, kostrikov2018discriminator}. %the demand for sampling complexity of interactions makes the AIL method hard to be implemented. %Some works have been proposed to reduce the number of interactions by leveraging model-based technique \cite{baram2017end} or off-policy approach \cite{sasaki2018sample}, however, 
However, the reinforcement loop inside AIL method has a risk of driving the learned policy to visit unsafe or undefined states-space during training. As such, in this work, we attend  %interested in 
the scenario where the imitated policy is NOT allowed to interact with environments or access information from the expert policy in training phase.

For the scenarios where interactions are not accessible, a common approach is still the Behavior Cloning \cite{pomerleau1989alvinn,schaal1999imitation}, which imitates the expert by approximating the conditional distribution of actions over states in a supervised learning fashion.
With sufficient demonstrations collected from the expert, the BC methods have successfully found its %capability of handling complicated applications, 
wide applications in autonomous driving \cite{bojarski2016end} and robot locomotion \cite{torabi2018behavioral}. Nevertheless, the robustness of BC is not guaranteed because of the compounding errors caused by covariate shift issue \cite{ross2011reduction}. 

%the imitated policy is trained based on single time-step state-action pair. Therefore, the BC based learned policy is susceptible to compounding errors caused by covariate shift \cite{ross2011reduction}. 

\begin{figure*}[h]
\centering
\includegraphics[width=0.95\textwidth]{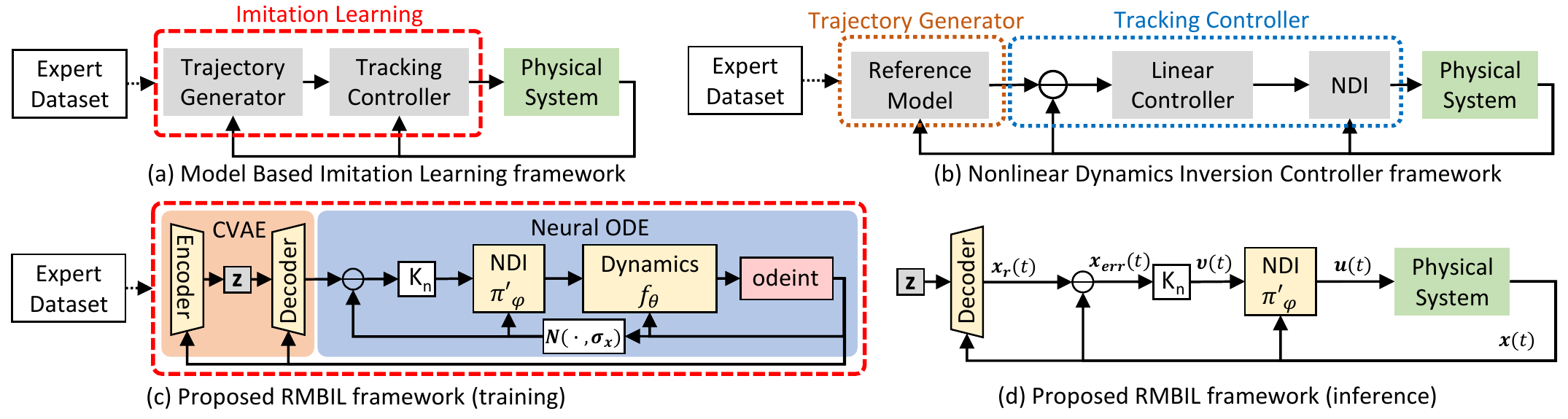}
\setlength{\abovecaptionskip}{-2pt}
\setlength{\belowcaptionskip}{-16pt}
\caption{Concepts behind the proposed RMBIL framework. Green block represents physical environment and yellow block represents the neural network. Red dash block indicates the target of the imitation learning. (a) Typical framework for Model-Based Imitation Learning (MBIL). (b) Classical block diagram for Nonlinear Dynamics Inversion (NDI) Controller framework, where the reference model (RM) is equivalent to the Trajectory Generator block in MBIL. A 
linear Controller and NDI block are formulated together as the Tracking Controller. (c) \& (d) Outline of the proposed RMBIL framework at training and inference phase, respectively, \textcolor{black}{Where the odeint block is a third-party numerical ode integrator.}  }% The RM block in NDI methodology is replaced by a Conditional Variational Autoencoder(CVAE), which is conditioned on the previous state. The $\text{K}_n$ block is a user defined feedback gain. The $\mathcal{N}$ block is the noise injection function. The pink block could be any modern ODE solver\footnotemark. Both the NDI controller and system dynamics are trained via closed-loop Neural ODE. (d) Pipeline for the trained RMBIL at inference. Only the decoder of the learned CVAE and the learned NDI controller are used for imitating the expert's behaviors. }

\label{fig-concepts-comparsion}
\vspace{-6pt}
\end{figure*}

%\vspace{-28pt}

Some efforts have been made to address the compounding errors issue under BC framework. Ross and Bagnell proposed DAgger \cite{ross2011reduction} that enables the expert policy to correct the behaviors of imitated policy during training. \textcolor{black}{Mahler and Goldberg \cite{laskey2017dart}} introduced Disturbances for Augmenting Robot Trajectory (DART) that applies the expert policy to generate sub-optimal demonstrations. Torabi et al. presented BCO \cite{torabi2018behavioral} with an inverse dynamics model for inferring actions from observations through environment exploration. % introduced %using noise injection in order to cover larger state distribution in which the imitated policy may encounter during execution. 
%Torabi et al. presented BCO \cite{torabi2018behavioral} with an inverse dynamics model for inferring actions from observations through environment exploration. %Another approach 
%Englert et al. treated the BC as a model-based probabilistic trajectory matching problem. 
%and iteratively updated the gradient-based imitation policy 
%via trajectory rollout generated from the learned Gaussian Process (GP) forward dynamics \cite{englert2013model}. 
Nevertheless, all these algorithms require interactions with either the environment or the expert policy during training, which is against our problem settings. 

%To satisfy the problem settings means that all training data is collected from the expert's demonstrations only. Hence, 
In order to effectively utilize the embedded physical information from the finite expert demonstrations, we choose the model-based scheme rather than BC like or model-free approaches. However, performance of model-based approach is directly influenced by accuracy of the learned system dynamics, as well as stability and robustness of learned controller. In the deep learning field, such properties are hard to be verified and guaranteed. Therefore, we borrow concepts and definitions from the nonlinear control theories so as to analyze the proposed framework. %where several algorithms have been proven and implemented in complex real-world systems, such as helicopter attitude control\cite{sieberling2010robust} and jet fighter aerobatics control \cite{bosworth2010success}.

As depicted by Osa et al. \cite{osa2018algorithmic}, the model-based imitation learning (MBIL) problem could be considered as a typical closed-loop tracking control problem, which is composed of system dynamics, tracking controller, and trajectory generator (shown in Fig.\ref{fig-concepts-comparsion}-(a)). In traditional nonlinear control field, such tracking control problem can be transformed into an equivalent linear system through 
Nonlinear Dynamics Inversion (NDI) \cite{snell1992nonlinear, enns1994dynamic}) algorithm (shown in Fig.\ref{fig-concepts-comparsion}-(b)).
%nonlinearities cancellation (shown in Fig.\ref{fig-concepts-comparsion}-(b)). This concept is so-called Nonlinear Dynamics Inversion (NDI) \cite{snell1992nonlinear, enns1994dynamic}). %also known as Feedback Linearization \cite{krener1983linearization, mellinger2011minimum}. 
As a result, the transformed linear system can be flexibly controlled by a linear controller. % by a linear controller. 
By applying NDI concept to MBIL problem, we could %take the exiting analysis tools from 
adopt some nonlinear control methodology, such as stability and robustness analysis \cite{enns1994dynamic}. However, %in the case of dynamics model mismatch, the performance of NDI may degrade and unstable situations may occur. In other words, 
the stability of NDI is guaranteed if the dynamics model matches the physical system \cite{sieberling2010robust}.

The recent Neural ODE research \cite{chen2018neural} presents a new framework to learn a precise multi-steps continuous system dynamics with irregular time-series data. %Based on this neural network model, 
With this seminal theory, Rubanova et al. \cite{rubanova2019latent} and Zhong et al. \cite{zhong2019symplectic} further proposed some improvements in their works.
%propose ODE-RNNs that generalizes RNNs to have continuous-time hidden dynamics by Neural ODE. %The experimental results show ODE-RNNs could handle complicated long-term dynamics with irregularly-sampled data. But they only focus on the autonomous systems. 
%Zhonget al. \cite{zhong2019symplectic} introduce Symplectic ODENet, which provides a systematic way to incorporate prior knowledge of Hamiltonian dynamics with control signal into the Neural ODE model. 
%However, their framework assumes the external inputs are constant during training and are evaluated only on simple low-dimensional environments, such as Pendulum, Acrobot, etc. 
As one key part of this work, we advocate a Neural ODE based purely data-driven approach for learning a multi-steps continuous actuated dynamics by applying the zero-order-hold to the control inputs. 
%The experiments on Mujoco \cite{todorov2012mujoco} tasks via OpenAI \cite{brockman2016openai} show the state-of-art performance with the test errors lower than 0.01 for 200 steps forward rollouts (please refer Appendix-A).  
Based on the precise learned dynamics, we propose the Robust Model-Based Imitation Learning (RMBIL) framework, which formulates MBIL as an end-to-end differentiable nonlinear tracking control problem via NDI algorithm, as illustrated in Fig.\ref{fig-concepts-comparsion}-(c), where the system dynamics and NDI control policy are trained with %within
a closed-loop Neural ODE model. On the other hand, the trajectory generator block in MBIL methodology is equivalent to the Reference Model (RM) in NDI framework. The RM module is usually a hand-designed kinodynamics trajectory approximator which mimics the expert's behaviors via a high-order polynomial function. Inspired by recent works in trajectories predictions \cite{walker2016uncertain, babaeizadeh2017stochastic, felsen2018will}, we further adopt the Conditional Variational Autoencoder (CVAE) \cite{sohn2015learning,kingma2013auto}, conditioned on the previous state, to replace the hand-designed RM block. At inference, only the learned NDI controller and the decoder from the trained CVAE are used for imitating expert's trajectories, as shown in Fig.\ref{fig-concepts-comparsion}-(d). In addition, to increase robustness of the proposed framework, we take the concept of the sliding surface from the Sliding Mode Control (SMC), which is a traditional nonlinear robust control algorithm \cite{slotine1991applied}. Empirically, we find that the sliding surface could be approximated by injecting noise to the system state during the training of the NDI controller with a closed-loop Neural ODE model. As a result, RMBIL obtains an approximated 30\% performance increase over BC, and is competitive to GAIL approach, in the disturbed environments.

%% file: sections/preliminary.tex
\vspace{-4pt}
\section{PRELIMINARY}
\vspace{-2pt}
Given expert demonstrations $\mathcal{D}\!=\!\{\mathcal{D}^k\}\!^N_{k=1}\!=\!\{\bm{\xi}^k_T, \bm{s}^k\}^N_{k=1}$, where $\bm{\xi}^k_T = (\bm{x}^k_{e,t_i}, \bm{u}^k_{e,t_i})_{i=0}^{T\!-\!1}$ is a finite sequence of state-action pairs for $T \in \mathbb{N}$ samples, $\bm{x}^k_{e,t_i}$ is sampled from state trajectory $\bm{x}(t) \in \mathbb{R}^n$, $\bm{u}^k_{e,t_i}$ is sampled from policy outputs $\bm{u}(t) \in \mathbb{R}^m$, and $\bm{s}^k$ is a context vector that represents the initial state $\bm{x}^k(0)$ of sequence $k$. A common method to solve MBIL is to explicitly train a discrete forward dynamics:
$
    \bm{x}^k_{t_{i\!+\!1}} = f_{\theta}(\bm{x}^k_{t_i}, \bm{u}^k_{t_i})
$, as well as a tracking controller:
$
    \bm{u}^k_{t_i} = \pi_\phi(\bm{x}^k_{r,t_{i\!+\!1}}, \bm{x}^k_{t_i})
$, where the reference state $\bm{x}^k_{r,t_i}$ is provided by a trajectory generator:
$
\bm{x}^k_{r,t_{i\!+\!1}} = \pi_\psi(\bm{x}^k_{t_i})
$, for all $N$ tasks.
\iffalse 
The whole process is shown in Fig.\ref{fig-concepts-comparsion}-(a). Therefore, the objective of MBIL is to minimize the difference between the expert's trajectory and imitated trajectory with respect to the given demonstration $\mathcal{D}^k=\{\bm{\xi}^k_T, \bm{s}^k\}$ across $N$ tasks:
\begin{equation}
    \underset{\psi,\phi}{\text{minimize}} \sum_{k=1}^{N}\sum_{i=0}^{T\!-\!1} \ell(\bm{x}^k_{e,t_i}, \bm{x}^k_{t_i})
\end{equation}
where $\ell(.)$ is a L2-norm $\ell(a,b) = (a-b)^2$. 
\fi
Since all models should be independent of $s^k$, without losing the generality, we assume $N=1$ for the following derivation in order to remove superscript $k$ for simplicity.
\vspace{-2pt}
\subsection{Nominal Control: Nonlinear Dynamics Inversion}
\vspace{-2pt}
The main concept behind NDI is to cancel nonlinearities of a system via input-output linearization \cite{slotine1991applied}. To %mathematically 
review the theory of NDI, consider a continuous input-affine system defined as:
\vspace{-8pt}
\begin{align}
    \label{eqn-continuous-dynx}
    \bm{\dot{x}}(t) & = \bm{a}(\bm{x}(t)) + \bm{G}(\bm{x}(t))\bm{u}(t) \\
    \label{eqn-observation}
    & \bm{y}(t) = \bm{h}(\bm{x}(t))
\end{align}
where $\bm{y}(t) \!\in\! \mathbb{R}^k$ is the output vector, $\bm{a}\!:\! \mathbb{R}^n \!\rightarrow\! \mathbb{R}^n$ and $\bm{h}\!:\! \mathbb{R}^n \!\rightarrow\! \mathbb{R}^k$ are smooth vector fields, and $\bm{G} \!\in\! \mathbb{R}^{n\times m}$ is a matrix whose columns are smooth vector fileds. To find the explicit relation between the outputs $\bm{y}(t)$ and the control inputs $\bm{u}(t)$, Eq. \eqref{eqn-observation} will be differentiated repeatedly (for simplicity, we denote $\bm{x}(t)\! = \!\bm{x}$).
\begin{equation}
    \label{eqn-NDI-differentation}
    \bm{\dot{y}} = \frac{d\bm{h}(\bm{x})}{dt} =  
    \nabla \bm{h}(\bm{x})\bm{\dot{x}}(t) =
    \nabla \bm{h}(\bm{x})\bm{a}(\bm{x}) + \nabla \bm{h}(\bm{x})\bm{G}(\bm{x})\bm{u}
\end{equation}
%To compute the NDI control inputs $\bm{u}_{ndi}$ so as to enable the system dynamics (Eq. \eqref{eqn-continuous-dynx}) 
where $\nabla$ is the gradient operator. If the term $\nabla\!\bm{h}(\bm{\!x\!})\bm{G}(\bm{\!x\!})$ in Eq.\eqref{eqn-NDI-differentation} is not zero, then the input-output relation is found. To achieve desired outputs $\bm{y}_{des}$, Eq.\eqref{eqn-NDI-differentation} is reformulated as:
\begin{equation}
    \label{eqn-NDI}
    \bm{u}_{ndi} = (\nabla \bm{h}(\bm{x})\bm{G}(\bm{x}))^{-\!1}[\bm{\nu}-\nabla \bm{h}(\bm{x})\bm{a}(\bm{x})]
\end{equation}
where $\bm{\nu} = \bm{\dot{y}}_{des}$ is a virtual input that is tracked by the derivative of output $\bm{\dot{y}}$. Hence, by controlling $\bm{\nu}$, the corresponding $\bm{u}_{ndi}$ in Eq.\eqref{eqn-NDI} would take $\bm{y}$ in Eq.\eqref{eqn-observation} to the desired output $\bm{y}_{des}$ through Eq.\eqref{eqn-continuous-dynx}. A common approach to obtain suitable $\bm{\nu}$ is using a linear proportional feedback controller: 
\begin{equation}
    \label{eqn-NDI-LPC}
    \bm{\nu} = \bm{K}_{n}(\bm{y}_{des}-\bm{y})
\end{equation}
where $\bm{K}_{n}$ is a $k\!\times\!k$ diagonal matrix \textcolor{black}{with a manually chosen gain value}. Eq.\eqref{eqn-NDI} will hold if $[\nabla \bm{h}(\bm{x})\bm{G}(\bm{x})]$ is invertible. If $\bm{G}(\bm{x})$ is a non-square matrix, then the pseudo-inverse method may be applied. In addition, we assume our system is fully-observable and $\bm{y} = \bm{x}$, which are conventions under MDP assumption in IL field \cite{ho2016generative,
wang2017robust, torabi2018behavioral}. Therefore, $\nabla \bm{h}(\bm{x})$ is an identity matrix, and Eq.\eqref{eqn-NDI} could be rewritten as follows:
\begin{equation}
    \label{eqn-NDI-tracking}
    \bm{u}_{ndi}(t) = \bm{G}^{-\!1}(\bm{x}(t))[\bm{\nu}(t)-\bm{a}(\bm{x}(t))]
\end{equation}
where $\bm{\nu}(t)$ is a function of $\bm{y}_{des}(t)$ and $\bm{y}(t)$ according to Eq.\eqref{eqn-NDI-LPC}. By applying Eq.\eqref{eqn-NDI-tracking} to the physical platform, the whole system becomes linear and can be flexibly controlled by Eq.\eqref{eqn-NDI-LPC}. However, stability of a NDI controller is guaranteed if the dynamics model (Eq.\eqref{eqn-continuous-dynx} and \eqref{eqn-observation}) matches the physical platform \cite{sieberling2010robust}. 
\vspace{-4pt}
\subsection{Ancillary Control: Sliding Mode Control}
\vspace{-2pt}
\label{subsec-SMC}
%Inspired by the recent work \cite{lopez2018robust} that proposes using the adaptive sliding control technique, which is a modified version of SMC, as an ancillary controller to increase the robustness of the nominal path planning, we use the SMC to assist the nominal NDI controller (as Eq.\eqref{eqn-NDI-LPC} and Eq.\eqref{eqn-NDI-tracking} in handling environment uncertainties. 
The idea behind SMC is to design a nonlinear controller to force the disturbed system to slide along the desired trajectory. As defined in \cite{slotine1983tracking}, the SMC scheme consists of two steps: (1) Find a sliding surface $\mathcal{S}$ such that the system will stay confined within it when the system trajectory reaches the surface within the finite time. In our case $\mathcal{S} = \bm{x}_{des}(t)$, where $\bm{x}_{des}(t) = \bm{y}_{des}(t)$ for the fully-observable system. (2) Design a switching function $\sigma(\bm{x}(t))$, which represents the distance between the current state and sliding surface, for the feedback control law $\bm{u}_{rn}$ to lead the system trajectory to intersect and stay on the $\mathcal{S}$. In this work, we define $\sigma(\bm{x}(t)) = \bm{x}_{err}(t) = \bm{x}_{des}(t)-\bm{x}(t)$ to satisfy the condition that $\sigma(\bm{x}(t)) = 0$ when $\bm{x}(t)=\mathcal{S}$.

The sufficient condition \cite{slotine1983tracking} for the global stability of SMC is $\bm{\sigma}^\intercal(\bm{x}(t))\bm{\dot{\sigma}}(\bm{x}(t))\!<\!0$. To satisfy the condition and dynamics model (Eq.\eqref{eqn-continuous-dynx}), the resulting nonlinear robust control policy $\bm{u}_{rn}$ is designed as follows (Denote $\bm{x}(t) =\bm{x}$):
\vspace{-4pt}
\begin{align}
    \label{eqn-SMC-NDI}
    \bm{u}_{rn} & \!=\! \bm{G}^{-\!1}(\bm{x})[\bm{\nu}-\bm{a}(\bm{x})+\bm{K}_s\!\cdot\!sgn(\bm{\sigma}(\bm{x}))] \\
    & \!=\! \bm{G}^{-\!1}(\bm{x})[\bm{\nu}\!\!-\!\!\bm{a}(\bm{x})]\!+\!\bm{G}^{-\!1}(\bm{x})\!\cdot\!\bm{K}_{s}\!\cdot\!sgn(\bm{\sigma}(\bm{x})) \\ 
    %& = \bm{u}_{ndi} + \bm{G}^{-\!1}(\bm{x})[\bm{K}_{s}\!\cdot\!sgn(\bm{\sigma}(\bm{x}))] \\
    \label{eqn-SMC}
    & \!=\! \bm{u}_{ndi} + \bm{u}_{smc}
\end{align}
where $\bm{K}_s$ is a positive definite $n\times n$ matrix. Therefore, by adding an additional feedback term $\bm{u}_{smc}$, which is dependent on the switching function $\bm{\sigma}(\bm{x})$, to the nominal NDI controller $\bm{u}_{ndi}$, we may obtain a robust NDI control law $\bm{u}_{rn}$. In addition, Eq.\eqref{eqn-SMC} also implies that $\bm{u}_{rn}$ would be identical to $\bm{u}_{ndi}$ when $\bm{x}_{\!err}\!=\!0$. %Please refer to Appendix for the derivation details of Eq.\eqref{eqn-SMC-NDI}.

%% file: sections/methodology.tex
\newcommand{\bp}[1]{\textcolor{blue}{#1}}
\section{METHODOLOGY}
\vspace{-2pt}
%\begin{comment}
\begin{figure}[t!]
\begin{center}
\centerline{\includegraphics[width=\columnwidth]{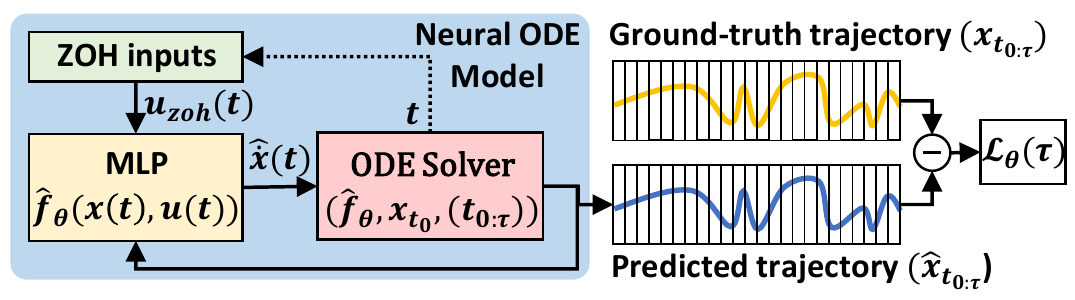}}
\caption{Actuated Neural ODE Model. Yellow block is a neural network for computing the time derivative of actuated dynamics. Green block is a continuous function $\bm{u}_\text{zoh}(\!t\!)$ approximated by ZOH method based on discrete input sequence $\bm{u}_{e,t_i}$. Although the internal state $\bm{x}(\!t\!)$ is continuous, the predicted outputs are discrete with regard to the specified time sequence $t_{0:\tau}$. Hence, the loss can be computed based on both discrete trajectories. }
% $\bm{x}_{t_{0:\tau}}$, 
% $\hat{\bm{x}}_{t_{0:\tau}}$.
\label{fig-actuated-node}
\end{center}
\vskip -0.2in
\end{figure}
%\vspace{-12pt}
%\end{comment}

\subsection{Multi-steps Actuated Dynamics using Neural ODE}
\vspace{-2pt}
Instead of learning a discrete dynamics, we are interested in a continuous-time dynamics with control inputs that can be formulated as: $\bm{\dot{x}}(t) = f_{\theta}(\bm{x}(t), \bm{u}(t))$. To approximate such differential dynamical systems with a neural network $\hat{f}_{\theta}$, we adopt the Neural ODE model proposed by Chen et al. \cite{chen2018neural}, which solves the initial-value problem (Eq.\eqref{eqn-IVP}) in a supervised learning fashion by backpropagating through a black-box ODE solver using adjoint sensitivity method \cite{pontryagin1962mathematical}  for memory efficiency.
\vspace{-3pt}
\begin{equation}
    \label{eqn-IVP}
    \bm{\hat{x}}_{t_0}, \dots, \bm{\hat{x}}_{t_n} = \mathrm{ODESolver}(\hat{f}_{\theta}, \bm{x}_{t_0}, (t_0, \dots, t_n) )
\end{equation}
To update the dynamics with respect to weights $\theta$, the loss function $\mathcal{L}_{\theta}$ can be constructed as a L2-norm between the predicted state $\bm{\hat{x}}_{t_i}$ and the true state $\bm{x}_{e,t_i}\!\in\!\mathcal{D}$ for a certain time-horizon $\tau$, that is: $\mathcal{L}_{\theta}(\tau) = \parallel\! \bm{x}_{e,t_{i:\tau}}\!-\!\bm{\hat{x}}_{t_{i:\tau}}\!\parallel^2$. However, %the approximated dynamics $\hat{f}_{\theta}$ in 
Eq.\eqref{eqn-IVP}  %assumed as 
can only handle an autonomous system $\bm{\hat{\dot{x}}}(t)\!=\!\hat{f}_{\theta}(\bm{x}(t))$. In order to apply Neural ODE to an actuated dynamics, Rubanova et al.\cite{rubanova2019latent} proposed a RNN based framework to map the system into latent space and solve the latent autonomous dynamics, while Zhong et al.\cite{zhong2019symplectic} introduced an augmented dynamics by appending the control input to the system state under a strong assumption that the control signal must stay constant over the time-horizon $\tau$. 

In this work, we propose a more intuitive and general method to handle the actuated dynamics. We construct a continuous input function $\bm{u}_\mathrm{zoh}(t)$ based on the sampled control signal $\bm{u}_{e,t_i}\!\in \!\mathcal{D}$ by zero-order-hold (ZOH). For each internal integration time $t_s$ inside Neural ODE, the actuated dynamics could always obtain the corresponding control signal by accessing the input function $\bm{u}_{e,t_s}\leftarrow\bm{u}_\mathrm{zoh}(t\!=\!t_s)$, as illustrated in Fig.~\ref{fig-actuated-node}. %, please refer to Appendix for graph illustration. %By implementing this method, 
As such, the Neural ODE (Eq.\eqref{eqn-IVP}) could be directly applied to an arbitrary dynamic system with external controls without any additional assumption or constraint. Once $\mathcal{L}_{\theta}(\tau) \rightarrow 0$ with $\tau \gg \Delta t$, where $\Delta t$ is the integration step of ODE solver, we obtain a precise multi-steps continuous actuated dynamics $\bm{\hat{\dot{x}}}(t)\!=\!\hat{f}_{\theta}(\bm{x}(t), \bm{u}(t))$ based on discrete expert demonstrations $\mathcal{D}$. %The experiments on complex Mujoco dynamics for the modified Neural ODE are shown in Appendix. 
%We apply the proposed framework on complex Mujoco dynamics with contact collision and evaluate the learned model by 200 steps forward prediction. The experiments are shown in Appendix. 
Unfortunately, compared to the baseline model such as multilayer perceptron (MLP), the integration solver inside Neural ODE becomes the bottleneck of inference time, which limits the learned dynamics to be integrated into classic model-based planning and control scheme such as Model Predictive Control (MPC). \cite{nagabandi2018neural, hafner2018learning, amos2018differentiable}.

\vspace{-4pt}
\subsection{Learning based NDI Controller via Neural ODE}
\vspace{-2pt}
To overcome the above inference time bottleneck caused by the ODE solver, we train a controller network via the learned dynamics with %within
the Neural ODE (Eq.\eqref{eqn-IVP}). As a result, only the learned controller will be adopted for controlling the physical system. Based on this concept, the Proposition \ref{prop-NDI} expresses how a control policy, parameterized by $\phi$, approximates the NDI formulation (Eq.\eqref{eqn-NDI-tracking}). % under certain assumptions.

\vspace{8pt}
\newtheorem{proposition}{\bf Proposition}[section]
\begin{proposition}
\label{prop-NDI}
{NDI Controller Training}
{\it \\Assume $\hat{f}_\theta(\bm{x}(t), \bm{u}(t))=\hat{a}_\theta(\bm{x}(t)) + \hat{G}_\theta(\bm{x}(t))\bm{u}(t)$, and suppose $\mathcal{L}_\phi(\tau)\approx0$, then $\bm{u}_{ndi}(t) \approx \hat{\pi}_\phi(\bm{\nu}(t), \bm{x}(t))$ if $\mathcal{L}_\theta(\tau)\approx0$ along trajectories, where $\tau \gg \Delta t$.}
\end{proposition}
\vspace{8pt}

\begin{figure*}[t!]
\begin{center}
\centerline{\includegraphics[width=0.85\textwidth]{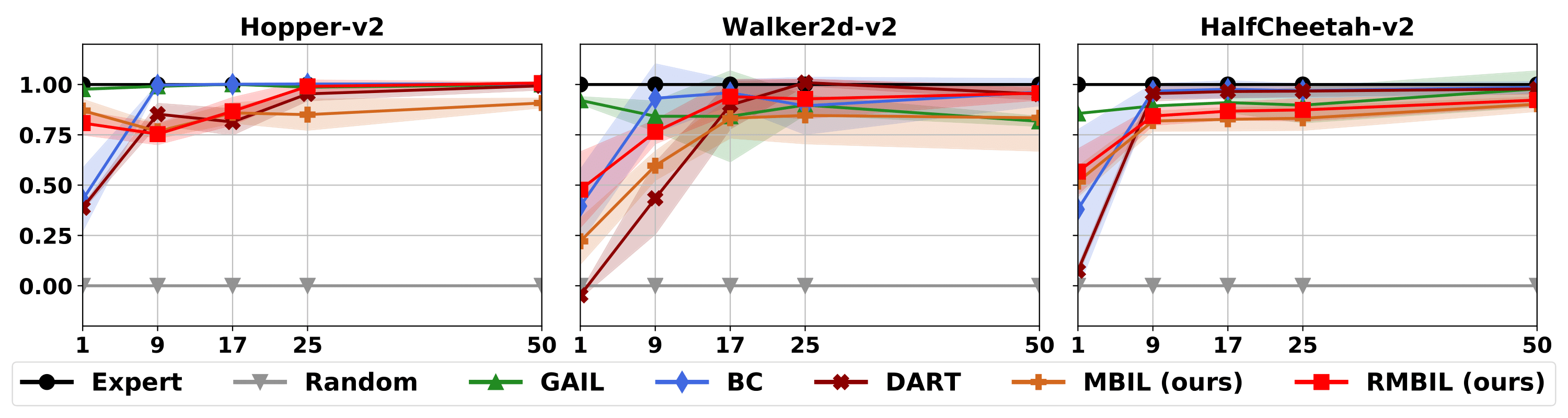}}
\setlength{\abovecaptionskip}{-2pt}
\setlength{\belowcaptionskip}{-20pt}
\caption{Performance comparison of proposed MBIL and RMBIL versus baselines with respect to the number of demonstrations. The x-axis is the number of expert demonstration used in the training, and y-axis is the normalized rewards (expert as one and random as zero). The shadow area beside each line represents its variance calculated based on 50 test trajectories. Note: GAIL needs environments interactions during training, while RMBIL does not.}
\label{fig-performance-comparsion}
\end{center}
\vskip -0.1in
\end{figure*}

 Proposition \ref{prop-NDI} implies that by minimizing the loss $\mathcal{L}_{\phi}(\tau)$, the NDI controller is learned with Neural ODE if the trained dynamics $\hat{f}_{\theta}$ is accurate and can be expressed as an affine system. To prove this proposition, we start from the loss function with respect to the controller weight $\phi$. 
\vspace{-4pt}
\begin{align}
    \label{eqn-NDI-loss}
    \mathcal{L}_{\phi}(\tau) & \!=\! \sum^{\tau}_{i=0}\{\bm{x}_{r,t_i}\!-\!\bm{\hat{x}}_{t_i}\}^2 \!=\! \{\bm{x}_{r,t_0}\!-\!\bm{\hat{x}}_{t_0}\}^2 \!+\!  \sum^{\tau}_{i=1}\{\bm{x}_{r,t_i}\!-\!\bm{\hat{x}}_{t_i}\}^2
\end{align}
\vspace{-4pt}
The first term at RHS of Eq.\eqref{eqn-NDI-loss} is zero since $\bm{x}_{t_0}$ is given as an initial condition. We let $\tau\!\!=\!\!1$ without loss of generality, and the results can be easily extended to multi-steps case.
\vspace{-2pt}
\begin{align}
    \mathcal{L}_{\phi}(\tau\!\!=\!\!1)
    &\!=\!\{\bm{x}_{r,t_1}\!\!-\!\bm{\hat{x}}_{t_1}\}\!^2
    \!=\! \{\bm{x}_r(t_1)\!-\!\bm{\hat{x}}(t_1)\}\!^2 \\ 
    %&\!=\! \{\bm{x}_r(t_1)\!-\!\int_{\Delta t}\bm{\hat{x}}(t_0)dt\,\}\!^2 \\
    \label{eqn-NDI-loss-dynx}
    & \!=\!\{\bm{x}_r(t_1)\!-\!\!\int_{\Delta t} \!\hat{f}_{\theta}[\bm{\hat{u}}(t_0), \bm{x}(t_0)]dt\,\}\!^2 
\end{align}
\vspace{-4pt}

The training goal of Neural ODE, which belongs to the supervised learning family, is to minimize the loss $\mathcal{L}_{\phi}$. According to Eq.\eqref{eqn-NDI-loss-dynx}, the loss will be the minimum if the first term equals to (or approximates) the second term as follows:
\vspace{-2pt}
\begin{equation}
    \label{eqn-NDI-min-cond}
    \bm{x}_r(t_1) = \int_{\Delta t} \!\hat{f}_{\theta}[\bm{\hat{u}}(t_0), \bm{x}(t_0)]dt 
\end{equation}
\vspace{-2pt}
In order to obtain the NDI formulation from Eq.\eqref{eqn-NDI-min-cond}, we introduce the first assumption that the equal sign still holds after applying the time derivative on both sides of Eq.\eqref{eqn-NDI-min-cond} if $\mathcal{L}_\theta(\tau)\approx0$ along trajectories, where $\tau \gg \Delta t$. In general, for arbitrary two functions whose value are identical at a certain point, the values of time derivative at the same point are not guaranteed to be equal. However, in Eq.\eqref{eqn-NDI-min-cond}, the state-action pairs' sequence data $\bm{\xi}_T$ used on both sides is from the same demonstrations $\mathcal{D}$. Therefore, if the training loss $\mathcal{L}_{\theta}$ of the learned dynamics approaches zero with $\tau \gg \Delta t$ (in practice, we stop the training as $\mathcal{L}_{\theta}\!<\! \epsilon$, where $\epsilon\!<\!0.01$), then both sides of Eq.\eqref{eqn-NDI-min-cond} represent the same system trajectory $\bm{x}(t)$. Hence, we could apply the time derivative operator to Eq.\eqref{eqn-NDI-min-cond} and the equal sign is still held as follows:
\vspace{-2pt}
\begin{align}
    \label{eqn-NDI-first-assume}
    \frac{d}{dt}\{\bm{x}_r(t_1)\} & = \frac{d}{dt}\{\int_{\Delta t} \!\hat{f}_{\theta}[\bm{\hat{u}}(t_0), \bm{x}(t_0)]dt\} \\
    \label{eqn-NDI-min-dot}
    \bm{\dot{x}}_r(t_1) & = \hat{f}_{\theta}[\bm{\hat{u}}(t_0), \bm{x}(t_0)]
\end{align}
To further %derivative 
derive Eq.\eqref{eqn-NDI-min-dot}, we introduce the second assumption that the true dynamics, where the proposed RMBIL is trying to mimic, is an input-affine system. This assumption is roughly satisfied with most physical controllable platforms. Under this assumption, Eq.\eqref{eqn-NDI-min-dot} can be reformulated as: 
\vspace{-4pt}
\begin{align}
    \label{eqn-NDI-assum-2}
    \bm{\dot{x}}_r(t_1) & = \hat{a}_\theta(\bm{x}(t_0)) + \hat{G}_\theta(\bm{x}(t_0))\hat{\bm{u}}(t_0) \\
    \label{eqn-NDI-with-pi}
    & = \hat{a}_\theta(\bm{x}(t_0)) + \hat{G}_\theta(\bm{x}(t_0))\hat{\pi}_{\phi}(\bm{\nu}(t_0),\bm{x}(t_0))
\end{align}
where $\bm{\nu}(t_0) = \bm{\nu}(\bm{x}(t_0),\bm{x}_r(t_1))$ from Eq.\eqref{eqn-NDI-LPC}. By substituting the controller network $\hat{\pi}_{\phi}$ for the expected control inputs $\hat{\bm{u}}(t)$, the relation to the NDI formulation emerges from Eq.\eqref{eqn-NDI-with-pi}. Since the two assumptions discussed above may not be fulfilled perfectly in practice, %such as $\mathcal{L}_{\theta} \rightarrow 0$ but $\mathcal{L}_{\theta} \neq 0$, 
we replace the equal sign with an approximation sign. In addition, we reformulate Eq.\eqref{eqn-NDI-with-pi} to bring $\hat{\pi}_{\phi}$ to the LHS:
\vspace{-2pt}
\begin{align}
    \label{eqn-NDI-appro}
    \hat{\pi}_{\phi}(\bm{\nu}(t_0),\bm{x}(t_0)) &\! \approx \! \hat{G}^{-\!1}_\theta(\bm{x}(t_0))[\bm{\dot{x}}_r(t_1)\!-\!\hat{a}_\theta(\bm{x}(t_0))] \\
    \label{eqn-NDI-final}
    \hat{\pi}_{\phi}(\bm{\nu}(t_0),\bm{x}(t_0))\! & \! \approx \bm{u}_{ndi}(t)
\end{align}
where Eq.\eqref{eqn-NDI-final} is derived from replacing the RHS of Eq.\eqref{eqn-NDI-appro} with Eq.\eqref{eqn-NDI-tracking}.
As a result, with Eq.\eqref{eqn-NDI-min-cond}, \eqref{eqn-NDI-first-assume} and \eqref{eqn-NDI-final}, Proposition \ref{prop-NDI} is validated. \textcolor{black}{We have to emphasize here that $\bm{u}_{ndi}(t)$ is only applied to the state space where we observed from the dataset $\mathcal{D}$}. Although the proposition indicates that a learned policy with %within
Neural ODE can be treated as a NDI controller, the robustness of the controller is not guaranteed \cite{sieberling2010robust}.

\iffalse
\begin{figure}[t!]
\begin{center}
\centerline{\includegraphics[width=\columnwidth]{figures/data-generation.png}}
\caption{Concept illustration of different data generation methods. Bold brown arrow is the mean and grey area is the distribution over expert trajectories. Brown dash arrow is a bias/correction trajectory from the expert at training. Blue dash arrow is a trajectory from the learner at training and Green arrow is a trajectory from the learner at inference. (a) The learner cannot recover when deviates the grey area. (b) The learner receives correction from the expert during training. (c) The expert generates sub-optimal trajectories to expand the gray area (d) The learner constructs a wider simulated distribution based on the learned dynamics.}
\label{fig-data-generation}
\end{center}
\vskip -0.2in
\end{figure}
\fi

\vspace{-4pt}
\subsection{Robustness Improvement through Noise Injection}
\vspace{-2pt}
\label{subsec-noise-injection}
To improve robustness of the learned NDI controller $\hat{\pi}_{\theta}$, we borrow the concept of SMC discussed in Section \ref{subsec-SMC}. Rather than building a hierarchical ancillary controller, we intend to design a robust NDI policy network $\hat{\pi}'_{\phi}$ which is end-to-end differentiable through Neural ODE. Inspired by DART \cite{laskey2017dart}, %that collects suboptimal demonstrations by injecting noise to the expert policy in order to cover wider trajectories distribution, 
we refine the trained NDI controller $\hat{\pi}_{\phi}$ by adding zero-mean gaussian noise to the internal state $\hat{\bm{x}}(t)$ within Neural ODE so as to construct the switching function $\bm{\sigma}$ (Eq.\eqref{eqn-SMC-NDI}). As stated with proposition \ref{prop-Robust}, the robustness of the refined controller $\hat{\pi}'_{\phi}$ would be improved because an ancillary SMC policy (Eq.\eqref{eqn-SMC}) is formulated automatically when $\mathcal{L}'_{\phi}$ (the training loss for $\hat{\pi}'_{\phi}$) approaching zero. Due to space limits, please refer to the link \footnote{\text{https://github.com/haochihlin/ICRA2021/blob/master/Appendix.pdf}} for the proof details. 

%of Proposition.\ref{prop-Robust}.

%The detailed proof process is ignored here due to space limitations(and we may  publish it online(weblink))

% \begin{wrapfigure}{r}{1.\linewidth}

% \begin{minipage}{1.\linewidth}

\vspace{6pt}
\begin{algorithm}[h]
   \caption{Dynamics and Controller Training}
   \label{alg:neural-ode-pipeline}
\begin{algorithmic}[1]
   \REQUIRE dataset $\mathcal{D}$, $\sigma_x$ for noise injection, $\tau$ for solver horizon, $\epsilon$ and $\epsilon_{r}$ as convergence indicator.
   
   // \texttt{Multi-steps dynamics training}
   \STATE Initialize model parameters $\theta$ and $\phi$ randomly
   \STATE Bypassing the controller model $\pi_{\phi}$
   \WHILE{$\mathcal{L}_{\theta}(\tau)>\epsilon$}
      \STATE Predict trajectories $\bm{\hat{x}}_{t_{i:\tau}}$ using Eq.\eqref{eqn-IVP}
      \STATE Compute loss $\mathcal{L}_{\theta}(\tau) = \parallel\! \bm{x}_{e,t_{i:\tau}}\!\!-\!\bm{\hat{x}}_{t_{i:\tau}}\!\parallel^2$
      \STATE Update the dynamics model $f_{\theta}$ via Neural ODE
   \ENDWHILE
   
   // \texttt{NDI controller training}
   \STATE Freeze trained dynamics $\!f_{\theta}$, enable controller $\pi_{\phi}\!$
   \WHILE{$\mathcal{L}_{\phi}(\tau)>\epsilon$}
      \STATE Predict trajectories $\bm{\hat{x}}_{t_{i:\tau}}$via Eq.\eqref{eqn-IVP} with trained $f_{\theta}$ 
      \STATE Update $\pi_{\phi}$ based on $\mathcal{L}_{\phi}(\tau) = \parallel\! \bm{x}_{e,t_{i:\tau}}\!\!-\!\bm{\hat{x}}_{t_{i:\tau}}\!\parallel^2$
   \ENDWHILE
   
   // \texttt{Controller robustness enhancement}
   \WHILE{$\mathcal{L}'_{\phi}(\tau)>\epsilon_r$}
      %\STATE Inject noise $\mathcal{N}(0,\sigma_x)$ into internal state $\hat{\bm{x}}(t)$ 
      \STATE Sample noised internal state $\hat{\bm{x}}'(t) \!\sim\! \mathcal{N}(\hat{\bm{x}}(t),\sigma_x)$
      \STATE Predict trajectories $\hat{\bm{x}}'_{t_{i:\tau}}$using trained $f_{\theta}$ and $\pi_{\phi}$
      \STATE Refine $\pi'_{\phi}$ based on $\mathcal{L}'_{\phi}(\tau) = \parallel\! \bm{x}_{e,t_{i:\tau}}\!\!-\!\hat{\bm{x}}'_{t_{i:\tau}}\!\parallel^2$
   \ENDWHILE
\end{algorithmic}
\end{algorithm}

\vspace{6pt}
\begin{proposition}
\label{prop-Robust} 
{Controller Robustness Improvement}
{\it
\\ Refine the learned controller $\hat{\pi}_{\phi}$ with noised injected state, sampled from Gaussian distribution, $\bm{x}'(t)\! \sim \! \mathcal{N}(\bm{x}(t),\sigma_x)$, under finite training epochs, then $\hat{\pi}'_{\phi}(\bm{\nu}'(t), \bm{x}'(t)) \approx \bm{u}_{rn}(t)$ when $\mathcal{L}'_\phi\rightarrow0$, where $\bm{\nu}'(t) = \bm{\nu}'(\bm{x}'(t),\bm{x}_r(t\!+\!\Delta t))$. }
\end{proposition}
\vspace{4pt}

%The proof of Proposition \ref{prop-Robust} is provided in Appendix. In addition, based on Proposition \ref{prop-Robust}, the concept of constructing an ancillary robust controller via noise injection can also be treated as an off-policy, model-based simulated data generation method.

\vspace{-4pt}
\subsection{Conditional VAE for Trajectory Generation}
\vspace{-2pt}
%Once the learnable robust NDI controller $\hat{\pi}'_{\phi}$ (Eq.\eqref{eqn-SMC-RHS}) is ready for inference, as illustrated in figure-\ref{fig-concepts-comparsion}, the RM block is required to generate a reference trajectory $\bm{x}_r(t)$ which will be tracked by the trained controller in order to mimic the expert demonstrations $\mathcal{D}^k$ under specific task $k$.%
%Unlike the learnable dynamics $\hat{f}_{\theta}$ or robust controller $\hat{\pi}'_{\phi}$, where the training objective is to approximate the true system globally, in other words, the learned $\hat{f}_{\theta}$ and $\hat{\pi}'_{\phi}$ should fit demonstrations $\mathcal{D}^k$ across all tasks, 
%the training objective for the trajectory generator in MBIL framework is to reproduce a narrow state distribution under specific task $k$ as described in Eq.\eqref{eqn-generator}. %With this purpose,

The training objective for the trajectory generator in MBIL framework is to predict the reference state given the past states.  \textcolor{black}{To support multi-task scenario}, inspired by \cite{babaeizadeh2017stochastic, wang2017robust} , we use CVAE \cite{sohn2015learning} to predict the future trajectory $\hat{\bm{x}}_r(t)$, but without the need for embedding multi-steps dynamics information because the learnable dynamics $\hat{f}_{\theta}$ and robust controller $\hat{\pi}'_{\phi}$ have carried on such information. At training, the generative mode of the proposed CVAE $q_{\psi_e}(\bm{z}|\bm{x}_{t_i}, \bm{c})$, parameterized by $\psi_e$, encodes the current state $\bm{x}_{t_i}$ to an embedding vector $\bm{z}$ conditioned on the variable $\bm{c}$, which in our case is the previous state $\bm{x}_{t_{i\!-\!1}}$. Given $\bm{z}$, the inference network $p_{\psi_d}(\bm{x}_{t_i}|\bm{z}, \bm{c})$, prarmeterized by $\psi_d$, decodes the current state $\bm{x}_{t_i}$ under the same condition variable $\bm{c}\!=\!\bm{x}_{t_{i\!-\!1}}$. Both network parameters ($\psi_e$ and $\psi_d$) are updated by minimizing the following loss:
\begin{equation}
    \setlength{\abovedisplayskip}{8pt}
    \setlength{\belowdisplayskip}{8pt}
    \label{eqn-cvae-loss}
    \begin{split}
    \mathcal{L}(\bm{x}_{t_i})\!= & - \mathbb{E}_{q_{\psi_e}(\bm{z}|\bm{x}_{t_i}, \bm{c})}[\text{log}\,p_{\psi_d}(\bm{x}_{t_i}|\bm{z}, \bm{c})] \\ 
    & + D_{K\!L}(q_{\psi_e}(\bm{z}|\bm{x}_{t_i}, \bm{c})||p(\bm{z|\bm{c}}))
    \end{split}
\end{equation}
where $D_{K\!L}$ is the Kullback-Leibler divergence between the approximated posterior and conditional prior $p(\bm{z}|\bm{c})$ which is assumed as the standard Gaussian distribution $\mathcal{N}(\bm{0},\bm{\text{I}})$. In addition, the first term on the RHS of Eq.\eqref{eqn-cvae-loss} represents the reconstruction loss.

For inference, by feeding the current system state $\bm{x}(t_i)$ as the context vector $\bm{c}$, the trained inference network $p_{\psi_d}(\bm{x}_{t_{i\!+\!1}}|\bm{z}, \bm{x}_{t_{i}})$ would predict the state at next time step $\bm{\hat{x}}(t_{i\!+\!1})$ which can be treated as the reference $\bm{x}_r(t_{i\!+\!1})$ for the tracking controller $\hat{\pi}'_{\phi}$, where the embedding vector $\bm{z}$ is sampled from the assumed prior $\mathcal{N}(\bm{0},\bm{\text{I}})$.

\vspace{-2pt}
\subsection{Procedure for Training and Inference}
\vspace{-2pt}
%In this subsection, we briefly summarize the implementation details for the proposed RMBIL framework in both training and inference phase. 
The training pipeline is composed of two main modules, the Neural ODE based dynamics-controller module and the CVAE based generator module. As described in Algorithm.\ref{alg:neural-ode-pipeline}, the dynamics-controller module is trained in three phases with %within
Neural ODE model. (1) Starting by training the dynamics model $\hat{f}_{\theta}$. (2) Once $\mathcal{L}_{\theta}(\tau) < \epsilon$, enable the control policy network $\hat{\pi}_{\phi}$ to learn a NDI controller. (3) When the loss $\mathcal{L}_{\phi}(\tau)<\epsilon$, start adding noise to the internal state $\hat{\bm{x}}(t)$ to obtain a robust controller $\hat{\pi}'_{\phi}$. The method to apply the noise injection inside Neural ODE is illustrated in Fig.\ref{fig-concepts-comparsion}-(c). In contrast, the training procedure of the generator module is straightforward, that is, minimizing the loss $\mathcal{L}_{\psi}$ defined in Eq.\eqref{eqn-cvae-loss} until both reconstruction loss and KL divergence converge. For inference, as illustrated in Fig.\ref{fig-concepts-comparsion}-(d), only the trained robust tracking controller $\hat{\pi}'_{\phi}$ %check symbols
and the trained inference network $p_{\psi_d}$ are used for driving the physical platform to mimic expert's behavior given initial state $\bm{x}_0$.

%As described in Algorithm. \ref{}, Dynamics-controller module is trained in three phases. Given expert demonstrations $\mathcal{}{D}$, (1) starts training the dynamics network $\hat{f}_{\theta}$ within Neural ODE bypassing the controller network $\hat{\pi}_{\phi}$. (2) Once the loss $\mathcal{L}_{\theta}$ is lower than a certain value $\epsilon$, freeze the trained dynamics model, then enable the controller network for training. (3) When the loss $\mathcal{L}_{\phi}$ converges, start adding Gaussian noise to the internal state $\bm{x}(t)$ in order to improve the robustness of the learned controller $\hat{\pi}'_{\phi}$. The training of the generator module is straightforward, that is minimizing the loss $\mathcal{L}_{\psi}$ defined in Eq.\eqref{eqn-cvae-loss} until both reconstruction loss and KL divergence converge. At inference, as illustrated in Fig.\ref{fig-concepts-comparsion}-(d), only the trained robust tracking control policy $\hat{\pi}'_{\phi}$ and the trained inference network $p_{\psi_d}$ are used for driving the physical platform to mimic expert's behavior given initial state $\bm{x}_0$ and optional context vector $\bm{s}$ (for multi-tasks case).

\begin{figure*}[t!]
\begin{center}
\centerline{\includegraphics[width=\textwidth]{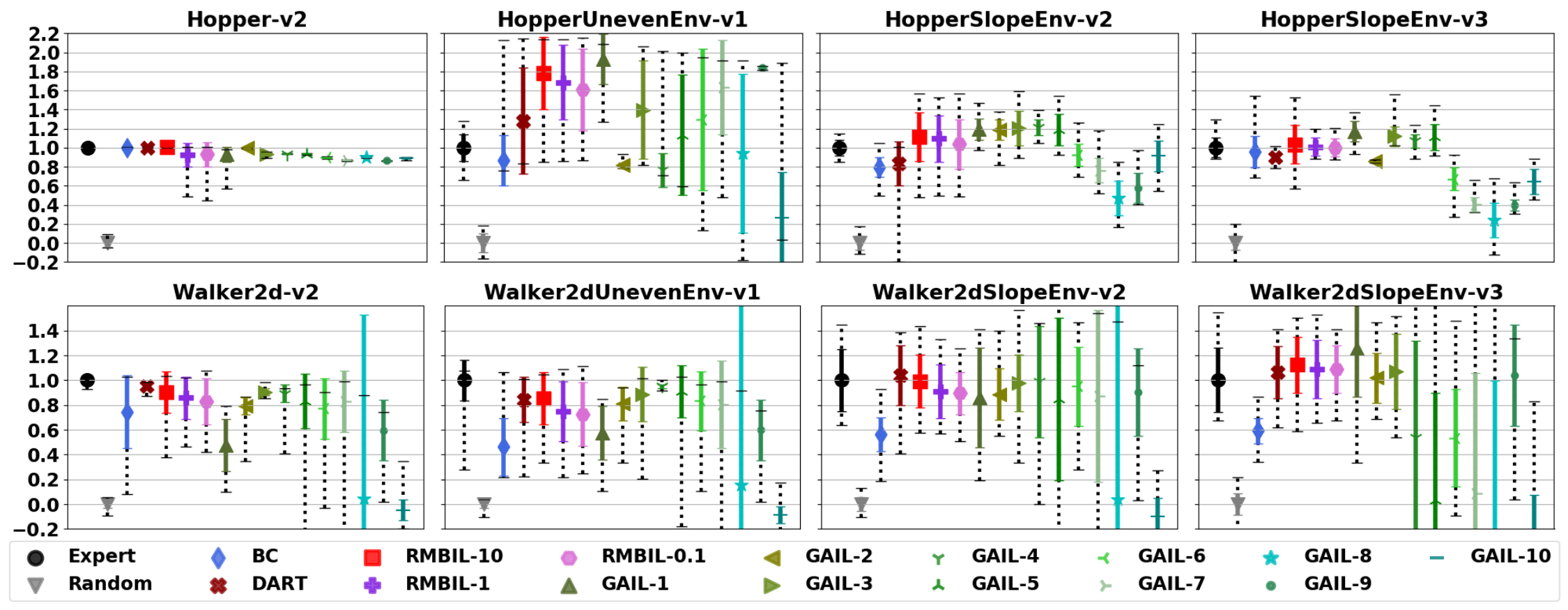}}
\setlength{\belowcaptionskip}{-28pt}
\caption{Robustness evaluation for RMBIL (with different feedback gains) versus baselines. The y-axis is the normalized rewards. For each method, we plot average (the marker), standard deviation (solid error bar) and minimum-maximum range (dash error bar) with respect to 50 test episodes. The number behind RMBIL stands for the value of the feedback gain used at inference. The number behind GAIL stands for the number of environment interactions used during training (x1000). Note: GAIL needs continuous  environments interactions during training, while RMBIL does not.}
\label{fig-robust-evaluation}
\end{center}
\vskip -0.2in
\end{figure*}
%\vspace{-2 pt}

%% file: sections/experiments.tex
\section{EXPERIMENTS}
\vspace{-2pt}
%In this section, we aim to answer the following questions with experiments. (1) Is the proposed RMBIL framework able to imitate the complicated behavior of continuous dynamics with collision? If so, how is the performance compared to the state-of-the-art algorithms? (2) In the environment with disturbances, how is the robustness of RMBIL compared to the existing methods? 

\subsection{Environment Setup} 
\vspace{-2pt}
We choose Hopper, Walker2d and HalfCheetah from OpenAI Gym \cite{brockman2016openai} under Mujoco \cite{todorov2012mujoco} physics engine as the simulation environments. To collect demonstrations, we use TRPO \cite{schulman2015trust} algorithm via stable-baselines \cite{stable-baselines} to train the expert policies. For each environment, we record the state-action pair sequence $\bm{\xi}_T^k$ with $T$ = 1000 steps for $k$ = 50 episodes under random initial state 
.%$\bm{x}_{e,t_0}^k$.

\vspace{-2pt}
\subsection{Baselines}
\vspace{-2pt}

We compare RMBIL against four baselines: BC, DART, GAIL and MBIL, where MBIL is a non-robust version of RMBIL (without noise injection). For BC method, a vanilla MLP is implemented to imitate the expert policy. For DART and GAIL methods, we slightly modify the official implementation to fit with our dataset format. For a fair comparison, the network size of the imitated policy for all methods are identical. The details of hyperparameters and network structure are listed in Table.\ref{tab-hyperparameters} . 

\input{tables/hyperparameters}

\vspace{-2pt}
\subsection{Performance Comparison}
\vspace{-2pt}
Given expert demonstrations, BC, MBIL, and RMBIL can be trained directly. For DART, we allow the algorithm to access the expert policies to generate sub-optimal trajectories during training. For GAIL, the training is unstable and requires interaction with the environment. Therefore, we train GAIL with $10^4$ interactions while saving the checkpoint every $10^3$ interactions, then choose the best one for comparison. At inference, for each method, we execute the trained policy for 50 episodes under random initial conditions defined by the environment, then compute the average and standard deviation. The normalized rewards against the number of demonstrations are shown in Fig. \ref{fig-performance-comparsion} by treating the performance of expert policy as one and random policy as zero. We can observe that RMBIL outperforms BC and DART in the case of extremely few demonstrations. Higher performance at low-data regimes implies that our method has better sample efficiency. With enough number of demonstrations, RMBIL can achieve similar performance to the expert policy, same as DART and BC. In contrast, the performance of GAIL is not directly dependent on demonstrations amount. %because the training of GAN is inherently unstable.
In addition, there is a performance gap between MBIL and RMBIL in high-data regimes (8\% for Hopper, 10\% for Walker2d). This phenomenon supports our discussion in Sec. \ref{subsec-noise-injection}., namely, the trained dynamics is not perfect, therefore, the learned controller should consider robustness in order to handle the model inaccuracies.

\input{tables/environment}
\vspace{-4pt}
\subsection{Robustness evaluation}
\vspace{-2pt}
In order to evaluate the robustness across different trained policies, we set up two types of environment disturbances for Hopper and Walker2D cases as described in TABLE.\ref{tab-environment}.
%(without HalfCheetah since its system is relatively stable compared to the other two), an uneven stair-like surface with random generated heights (between 0 and 20 mm) and a slope with certain angles (0.5 and 1 degrees). 
%Details of the environments are described in Appendix. 
Because we are interested in how the feedback gain $K_n$ of the linear controller inside RMBIL affects the robustness of the learned policy, we train RMBIL with $K_n\!=\!0.1$ based on 50 demonstrations. At inference, we compare the performance of learned robust NDI controller under different gain (0.1,1 and 10). The normalized average reward over 50 trajectories in Fig.\ref{fig-robust-evaluation} shows that RMBIL with high-gain obtains better mean than the ones with low-gain in both cases. This observation meets the linear control theory that robustness of the transformed linear system can be enhanced by increasing the feedback gain \cite{zhou1998essentials}. In addition, compared to DART, RMBIL has higher rewards mean in Hopper case and similar performance in Walker2d case. %However, the training of DART needs access to expert policy. %while ours do not need it. 
In contrast, the comparison of GAIL is hard to analyze since the performance of its imitated policy varying significantly with respect to the number of environments interactions. However, we could still roughly observe that the mean and variance of RMBIL is located on the average performance range of GAIL in both Walker2d and Hopper cases. %Under the same performance, GAIL is limited by the need for environment interaction during training.

%the other methods (even the expert policy) in Hopper case. For Walker2d case, RMBIL is competitive to DART and GAIL, moreover, our method has no need for either environment interaction or accessing the expert policy during training. 

On the other hand, through Walker2d cases in Fig.\ref{fig-robust-evaluation}, we could observe the covariate shift issue exists in the BC method, where the trained BC policy achieves the same rewards as the expert in the default environment, however, when encountering unknown disturbances, the performance of BC policy degrades dramatically. In comparison, since the proposed RMBIL is based on a precise multi-steps dynamics and a nonlinear controller with noise injection, the learned robust controller could avoid overfitting to the expert demonstrations and overcome the environment uncertainties at testing time. 
%As listed in Table.\ref{tab-rmbil-bc}, we could obtain around 28\% performance increase  (in terms of reward) in uneven Hopper environments and at least 45\% increase in uneven Walker2d environments by choosing RMBIL method. As such, we suggest that model-based would be a better solution than model-free in the context of our problem settings. The experiments for evaluating Neural ODE based multi-steps dynamics on Mujoco tasks are described in Appendix.

%% file: tables/hyperparameters.tex
\begin{table}[h]
\caption{Hyperparameters for RMBIL training}
\label{tab-hyperparameters}
\begin{center}
\begin{small}
\vspace{-12pt}
\setlength{\tabcolsep}{4pt}
\begin{tabular}{|cc|}
\hline
Hyperparameter & Value \\
\hline
No. of hidden neuron ($\theta$) & 800 \\
No. of hidden neuron ($\phi$ and $\psi$) & 320 \\
No. of hidden layers ($\theta$, $\phi$ and $\psi$) & 2 \\
activation functions ($\theta$, $\phi$ and $\psi$) & ELU \\
latent dimension $\bm{z}$ for $\psi$ & 5$\sim$10\\
learning rate ($\theta$) & 0.01 \\
learning rate ($\phi$ and $\psi$) & 0.001 \\
learning rate decay & 0.5 per 100 epochs \\
type of ode-solver & adams\\
absolute tolerance for ode-solver & 1e-4 \\
relative tolerance for ode-solver & 1e-4 \\
noise standard deviation $\sigma_{\bm{x}}$ & 0.25 \\
batch size & 2048  \\
\hline

\end{tabular}
\end{small}
\end{center}
\vskip -0.2in
\end{table}

%% file: tables/environment.tex
\begin{table}[h]
\caption{Robustness Environments Setting}
\label{tab-environment}
\begin{center}
\begin{small}
\vspace{-16pt}
\setlength{\tabcolsep}{4pt}
\begin{tabular}{|c|c|c|}
\hline
Case & UnevenEnv & SlopeEnv \\
\hline
Hopper & (v1) 1m span boxes with & (v2) 0.5 deg slope\\
       & random heights (0 $\sim$ 20mm) & (v3) 1.0 deg slope\\
\hline
Walker2d & (v1) 2m span boxes with & (v2) 0.5 deg slope\\
         & random heights (0 $\sim$ 20mm) & (v3) 1.0 deg slope\\
\hline
\end{tabular}
\end{small}
\end{center}
\vspace{-10pt}
\end{table}

%% file: sections/conclusion.tex
\vspace{-15pt}
\section{CONCLUSION}
\vspace{-6pt}
In this work, we presented RMBIL, a Neural ODE based approach for imitation learning without the need for access to expert policy or environment interaction during training.  %From the perspective of traditional nonlinear control theory, we propose a new differentiable model-based %imitation learning 
%IL framework, which has no need for environment interaction during training.
To the best of our knowledge, we are the first to study IL problem from the perspective of traditional nonlinear control theory with both theoretical and empirical supports. With the theoretical analysis, we prove that the learnable control network inside Neural ODE could approximate an NDI controller by minimizing the training loss. Experiments on complicated Mujco tasks show that RMBIL can achieve the same performance as the expert policy. In addition, for unstable systems, such as Hopper and Waker2d, with environmental disturbances, the performance of RMBIL is competitive to GAIL algorithm and outperforms BC method. Future works may incorporate other existing classic nonlinear control theories and explore multi-tasks applications.